\documentclass{article}

\usepackage{authblk}
\usepackage{lmodern}
\usepackage{url}
\usepackage{hyperref}
\usepackage{listings}
\usepackage{booktabs}
\usepackage{color}
\usepackage{amsmath}
\usepackage{natbib}
\usepackage{amssymb}
\usepackage{graphicx}
\usepackage[margin=1.25in]{geometry}
\usepackage{lastpage}
\usepackage[dvipsnames]{xcolor}

\definecolor{codegreen}{rgb}{0,0.6,0}
\definecolor{codegray}{rgb}{0.5,0.5,0.5}
\definecolor{codepurple}{rgb}{0.58,0,0.82}
\definecolor{backcolour}{rgb}{0.95,0.95,0.92}

\lstdefinestyle{mystyle}{
  backgroundcolor=\color{backcolour}, commentstyle=\color{codegreen},
  keywordstyle=\color{magenta},
  numberstyle=\tiny\color{codegray},
  stringstyle=\color{codepurple},
  basicstyle=\ttfamily\footnotesize,
  breakatwhitespace=false,         
  breaklines=true,                 
  captionpos=b,                    
  keepspaces=true,                 
  numbers=left,                    
  numbersep=5pt,                  
  showspaces=false,                
  showstringspaces=false,
  showtabs=false,                  
  tabsize=2
}
\lstdefinelanguage{json}{
  basicstyle=\ttfamily\footnotesize,
  numbers=left,
  numberstyle=\tiny\color{codegray},
  stepnumber=1,
  numbersep=5pt,
  showstringspaces=false,
  breaklines=true,
  frame=single,
  backgroundcolor=\color{backcolour},
  morestring=[b]",
  moredelim=[s][\color{codepurple}]{\"}{\"},
  morecomment=[l]{//},
  commentstyle=\color{gray}\ttfamily,
  morekeywords={true,false,null},
  keywordstyle=\color{Fuchsia},
  literate=
   *{:}{{{\color{Sepia}:}}}{1}
    {,}{{{\color{Sepia},}}}{1}
    {0}{{{\color{blue}0}}}{1}
    {1}{{{\color{blue}1}}}{1}
    {2}{{{\color{blue}2}}}{1}
    {3}{{{\color{blue}3}}}{1}
    {4}{{{\color{blue}4}}}{1}
    {5}{{{\color{blue}5}}}{1}
    {6}{{{\color{blue}6}}}{1}
    {7}{{{\color{blue}7}}}{1}
    {8}{{{\color{blue}8}}}{1}
    {9}{{{\color{blue}9}}}{1}
}

\author[1]{Marc González}
\author[2]{Rachid Guerraoui}
\author[3]{Rafael Pinot}
\author[2]{Geovani Rizk}
\author[2]{John Stephan\thanks{Corresponding author: \texttt{john.stephan@epfl.ch}.}}
\author[4]{François Taïani}

\affil[1]{Universitat Politècnica de Catalunya (UPC)}
\affil[2]{Ecole Polytechnique Fédérale de Lausanne (EPFL)}
\affil[3]{Sorbonne Université, Université Paris Cité, CNRS}
\affil[4]{Université de Rennes, Inria, CNRS, IRISA}
\affil[ ]{\textit{Authors are listed in alphabetical order.}}

\date{}

\begin{document}

\title{\textsc{ByzFL}: Research Framework for Robust Federated Learning}

\maketitle
\begin{abstract}
We present \textsc{ByzFL}, an open-source Python library for developing and benchmarking robust federated learning (FL) algorithms. \textsc{ByzFL} provides a unified and extensible framework that includes implementations of state-of-the-art robust aggregators, a suite of configurable attacks, and tools for simulating a variety of FL scenarios, including heterogeneous data distributions, multiple training algorithms, and adversarial threat models.
The library enables systematic experimentation via a single JSON-based configuration file and includes built-in utilities for result visualization.
Compatible with PyTorch tensors and NumPy arrays, \textsc{ByzFL} is designed to facilitate reproducible research and rapid prototyping of robust FL solutions. \textsc{ByzFL} is available at \url{https://byzfl.epfl.ch/}, with source code hosted on GitHub: \url{https://github.com/LPD-EPFL/byzfl}.
\end{abstract}

\section{Introduction}
Federated Learning (FL) has emerged as a promising paradigm for machine learning (ML) in settings where data privacy, sovereignty, and scalability are paramount. Rather than centralizing data on a single machine, FL enables multiple \textit{clients}, such as hospitals, financial institutions, or edge devices, to collaboratively train a shared global model.
This is achieved by transmitting model updates to a central trusted \textit{server} for aggregation, while keeping raw data local~\citep{NIPS2012_6aca9700, bertsekas2015parallel, abadi2016tensorflowlargescalemachinelearning, mcmahan17a, kairouzfl}.
Such a distributed approach is especially well-suited for domains like healthcare and finance~\citep{DBLP:conf/nips/TerrailACGHLMMM22}, where regulatory and ethical constraints often prohibit the sharing of sensitive information across institutions or jurisdictions. In addition to alleviating legal concerns, FL mitigates the computational bottlenecks associated with centralized training.
As modern ML models continue to scale to hundreds of billions or even trillions of parameters~\citep{chowdhery2022palmscalinglanguagemodeling, smith2022usingdeepspeedmegatrontrain}, distributing the training workload across clients becomes necessary. FL thus provides a scalable framework for building ML systems in sensitive and distributed environments.
\medskip

\begin{figure}[ht!]
    \centering
    \includegraphics[scale=0.15]{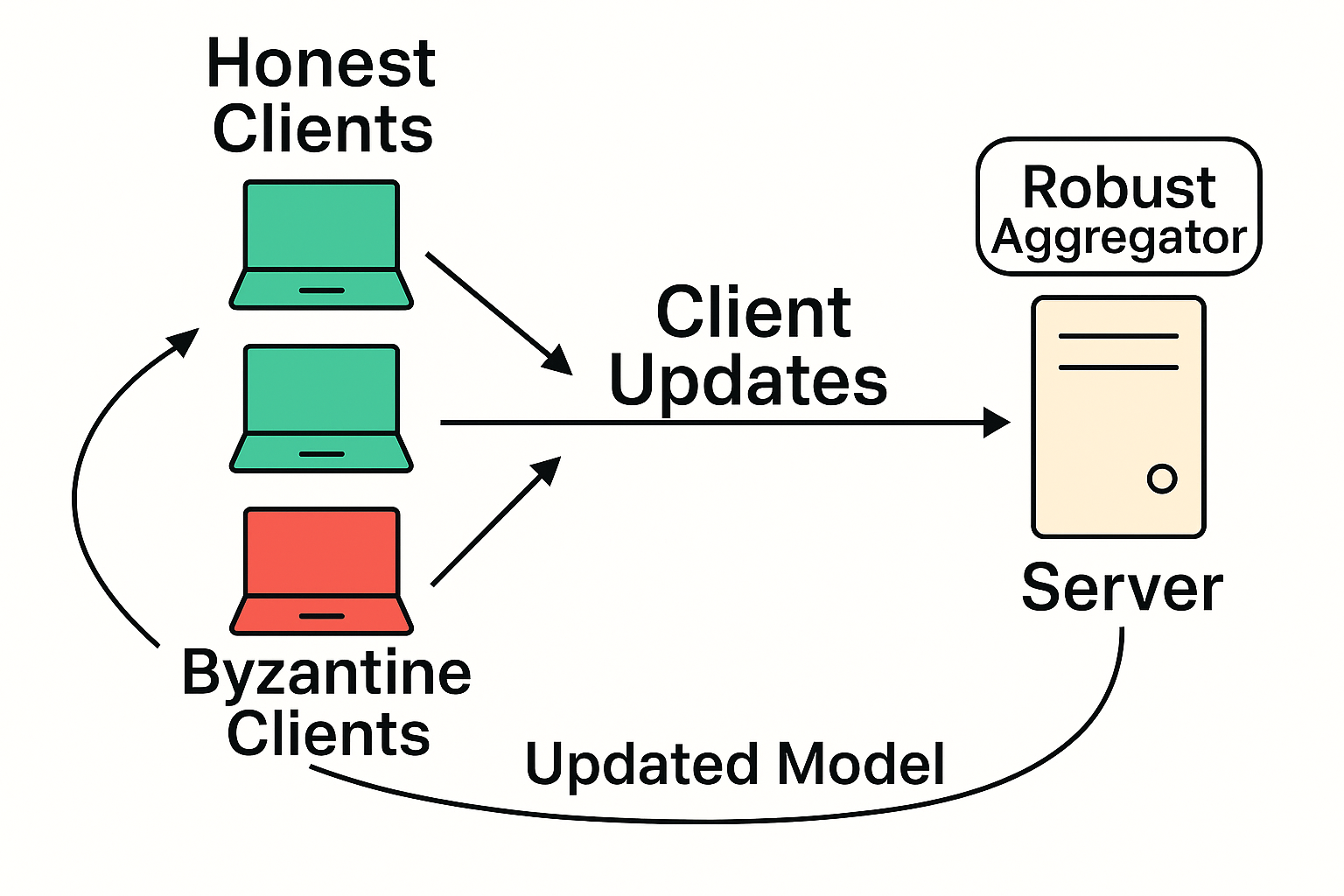}
    \vspace{-10pt}
    \caption{FL setup with server-side robust aggregation of client updates}
    \label{fig:rob_agg}
\end{figure}

However, FL also introduces a unique set of challenges. It is, in particular, vulnerable to faulty or adversarial clients. Such clients may deviate from the prescribed learning protocol due to hardware or software failures, corrupted local data~\citep{poisattacksBiggio12, universalMahloujifar2019}, or malicious intent~\citep{localFang2020}. These misbehaving clients are often referred to as \textit{Byzantine}, in reference to the classical fault model from distributed computing~\citep{lamport82}.
Standard FL algorithms offer no inherent protection against such adversaries~\citep{blanchard2017},  which poses significant risks for deployment in critical domains such as healthcare, finance, or public policy~\citep{Gernot16Datatrust, Bell21Replacing, Pi_2021, Colin22}.
Ensuring robustness to Byzantine behavior is therefore essential for the safe and responsible deployment of FL systems in practice.
Over the past decade, substantial progress has been made toward addressing robustness challenges in FL. Since the foundational works that formalized the problem~\citep{blanchard2017, yin2018byzantine}, a wide range of robust training algorithms have been proposed. These typically enhance the server-side update step by replacing naive averaging with more sophisticated robust aggregation schemes (see Figure~\ref{fig:rob_agg}).
These defenses range from coordinate-wise and gradient filtering techniques to geometric aggregation schemes. Collectively, these advances have led to a mature theoretical understanding of the guarantees and limitations of robust FL~\citep{allouah2023fixing, allouah2023robust, farhadkhani24FL}.

\paragraph{Contribution.}
Despite the rapid growth of research on Byzantine-resilient FL, the field still lacks a standardized and reproducible benchmarking framework for evaluating robustness under adversarial conditions.
Most experimental results are derived from custom implementations, each with different assumptions, threat models, and evaluation metrics. This makes it difficult to compare methods fairly or to assess the generality of robustness claims.
As a result, progress in robust FL remains hard to track empirically, and new defenses are often validated in narrow or overly specific settings.
To address this gap, we introduce \textsc{ByzFL}, an open-source Python library for designing, testing, and benchmarking robust FL algorithms.
\textsc{ByzFL} is built to lower the barrier to entry for both researchers and practitioners working on robustness in ML. It is fully compatible with PyTorch tensors~\citep{pytorch} and NumPy arrays~\citep{numpy}, and has been designed to be accessible to both beginners and experts.
The library serves as a reproducible testbed for systematic experimentation, and provides a unified interface for evaluating robust aggregation methods under diverse configurations.
\textsc{ByzFL} is available at \url{https://byzfl.epfl.ch/}, with source code hosted on GitHub: \url{https://github.com/LPD-EPFL/byzfl}.

\paragraph{Comparison with existing FL frameworks.}
Several open-source frameworks have been developed to support FL experimentation and deployment, including \textit{LEAF}~\citep{caldas2018leaf}, \textit{FedML}~\citep{he2020fedml}, \textit{Flower}~\citep{beutel2022flower}, \textit{TensorFlow Federated (TFF)}~\citep{tensorflowfederated2019}, and \textit{PySyft}~\citep{ryffel2018pysyft}.
These platforms provide valuable tools for simulating FL at scale, handling heterogeneous clients, or deploying real-world FL systems.
However, these frameworks are not specifically designed for evaluating robustness to adversarial clients. In particular, they do not natively support Byzantine attack simulation or systematic benchmarking of robust aggregation methods.
The work most closely related to ours is \textit{Blades}~\citep{li2024blades}, a recent Python benchmarking framework for robust FL. While Blades provides code to reproduce the experiments from its associated paper, it is not designed to be extended or built upon.
In contrast, \textsc{ByzFL} is constructed from the ground up to support continuous development and community contributions. It introduces a fully configurable and user-friendly benchmarking system controlled by a single JSON file, requiring no code changes to define entire experimental pipelines.
Beyond benchmarking, \textsc{ByzFL}'s modular architecture allows robust aggregation methods and attack models to be used independently of any training pipeline, operating directly on PyTorch tensors or NumPy arrays. This design makes \textsc{ByzFL} both a benchmarking suite and a flexible research toolkit, enabling easy integration into diverse workflows and fostering experimentation and innovation in robust FL.

\paragraph{Paper outline.} The remainder of this paper is organized as follows.
Section~\ref{aggregators} describes the robust defense mechanisms and adversarial attack strategies implemented in \textsc{ByzFL}.
Section~\ref{framework} outlines the overall architecture of the library, including its simulation framework for federated training pipelines and its benchmarking suite for evaluating robustness.
Finally, Section~\ref{conclusion} concludes the paper with a discussion of the library’s impact.

\section{Robust Aggregation and Adversarial Threat Models}\label{aggregators}
To effectively assess the robustness of the global model in the presence of adversarial clients, two key ingredients are required: (i) robust aggregation methods that can tolerate corrupted updates, and (ii) principled adversarial threat models for systematically evaluating their effectiveness.
In this section, we present how \textsc{ByzFL} enables both the implementation of robust aggregation defenses and their evaluation against diverse adversarial threat models.

\paragraph{Robust (pre-)aggregators.}
At the core of FL algorithms, averaging is the standard approach for aggregating model updates. It is used in \textit{FedAvg}~\citep{mcmahan17a}, \textit{Distributed Stochastic Gradient Descent (DSGD)}~\citep{bertsekas2015parallel}, and more advanced methods such as \textit{FedProx}~\citep{Li20FedOpt} and \textit{FedYogi}~\citep{reddi2021adaptive}. However, averaging is a linear operation and can be arbitrarily manipulated by even a single adversarial client~\citep{blanchard2017}. To address this vulnerability, a significant body of research has proposed replacing naive averaging with robust aggregators that preserve the integrity of the global model in the presence of malicious clients. These defenses have been shown to improve the robustness of FL both theoretically and empirically~\citep{farhadkhani22, allouah2023fixing, farhadkhani24FL}. \textsc{ByzFL} implements a wide range of state-of-the-art robust aggregators, including \textit{MultiKrum}~\citep{blanchard2017}, \textit{geometric median}~\citep{chen2017distributed}, \textit{coordinate-wise median} and \textit{trimmed mean}~\citep{yin2018byzantine}, \textit{mean around median}~\citep{meamed}, \textit{minimum diameter averaging}~\citep{brute_bulyan}, \textit{centered clipping}~\citep{karimireddy21a}, \textit{MoNNA}~\citep{farhadkhani23a}, \textit{smallest maximum eigenvalue averaging}~\citep{allouah2023trilemna}, and \textit{covariance-bound agnostic filter}~\citep{allouah2025trustworthyfederatedlearninguntrusted}. \medskip

In practice, however, the effectiveness of robust aggregators can be significantly hindered in settings with high data heterogeneity. When clients' data distributions are highly non-IID, even honest updates may diverge substantially. As a result, robust methods can mistakenly discard them as outliers, ultimately degrading convergence. To mitigate this, pre-aggregation serves as a complementary mechanism: it reduces the variability among honest updates before applying robust aggregation. This creates a hierarchical pipeline in which pre-aggregation acts as a normalization or denoising step, enhancing the downstream aggregator’s tolerance to both adversarial noise and benign heterogeneity. \textsc{ByzFL} implements four prominent pre‑aggregation techniques from the literature: \textit{nearest neighbor mixing (NNM)}~\citep{allouah2023fixing}, \textit{bucketing}~\citep{karimireddy2021byzantine}, \textit{static clipping}, and \textit{adaptive robust clipping (ARC)}~\citep{allouah2025adaptive}. Listing~\ref{lst:robg_agg} illustrates how to configure a robust aggregation pipeline in \textsc{ByzFL} using the \texttt{MultiKrum} aggregator pre-composed with the \texttt{NNM} pre-aggregator.
\medskip

Robust mean estimation can be viewed as a foundational building block of robust FL.
It corresponds to the simplified setting where each client contributes a single vector (e.g., a gradient), clients hold i.i.d. data, and the server performs a one-shot aggregation to estimate the mean of correct inputs, while defending against a fraction of adversarial inputs. Many robust aggregators used in FL are inspired by classical results from robust statistics.
In \textsc{ByzFL}, we deliberately decouple the implementation of robust aggregators from the FL pipeline (see Section~\ref{framework}). This modular design enables researchers and practitioners in robust statistics to leverage our tools beyond FL, wherever robust aggregation is needed.

\begin{lstlisting}[language=Python, caption=Robust aggregation in \textsc{ByzFL} using \texttt{MultiKrum} and \texttt{NNM}, label={lst:robg_agg},float]
import byzfl, numpy

# 3 input vectors
x = numpy.array([[1., 2., 3.], 
                 [4., 5., 6.],
                 [7., 8., 9.]])

# f is the number of Byzantine clients
f = 1

# Pre-aggregate using NNM and aggregate using MultiKrum
pre_agg = byzfl.NNM(f=f)
agg = byzfl.MultiKrum(f=f)
aggregated_vector = agg(pre_agg(x)
print("Aggregated vector:", aggregated_vector))
# Output: Aggregated vector: array([2.5 3.5 4.5])
\end{lstlisting}

\paragraph{Simulating adversarial behavior.}
In \textsc{ByzFL}, misbehaving clients are modeled through a suite of well-established attack strategies.
The library currently implements a range of canonical Byzantine attacks, including \textit{Sign Flipping} and \textit{Label Flipping}~\citep{allen2020byzantine}, \textit{Inner Product Manipulation (IPM)}~\citep{cong19fall}, and \textit{A Little Is Enough (ALIE)}~\citep{baruch19little}. Since both IPM and ALIE are governed by a tunable parameter $\tau \in \mathbb{R}$, referred to as the attack factor, \textsc{ByzFL} extends these attacks with adaptive variants, named \textit{Opt-IPM} and \textit{Opt-ALIE}. Unlike their standard counterparts that rely on a fixed $\tau$, these optimized versions dynamically adjust the attack factor based on the characteristics of the current input. Specifically, given a set of honest client updates, \textsc{ByzFL} performs a grid search over a predefined range of $\tau$ values to identify the optimal attack factor $\tau_{\text{opt}}$ that maximizes adversarial impact. The optimization objective is to maximize the $\ell_2$-norm of the discrepancy between the mean of the honest inputs and the output of the aggregator.
\medskip

The attacks implemented within \textsc{ByzFL} span a spectrum of adversarial strategies, from untargeted random corruption to maliciously optimized gradient perturbations.
By simulating these diverse strategies, \textsc{ByzFL} enables systematic evaluation and stress-testing of robustness claims and supports the development of new aggregation defenses that explicitly account for adversarial behavior in FL.
As an example, Listing~\ref{lst:attack} illustrates how to execute the \texttt{SignFlipping} attack in \textsc{ByzFL} using NumPy arrays. For completeness, we demonstrate in Listing~\ref{lst:pipeline} (Appendix~\ref{app_code}) how the attack and robust aggregator modules of \textsc{ByzFL} can be used directly on PyTorch tensors.

\begin{lstlisting}[language=Python, caption=Executing the \texttt{SignFlipping} attack within \textsc{ByzFL}, label={lst:attack},float]
import byzfl, numpy

# 3 honest vectors
x = numpy.array([[1., 2., 3.], 
                 [4., 5., 6.],
                 [7., 8., 9.]])

# Execute attack
attack_vector = byzfl.SignFlipping(x)
print("Attack vector:", attack_vector)
# Output: Attack vector: array([-4. -5. -6.])
\end{lstlisting}

\section{Federated Simulation Framework}
\label{framework}
\begin{figure}[ht!]
    \centering
    \includegraphics[width=1\linewidth]{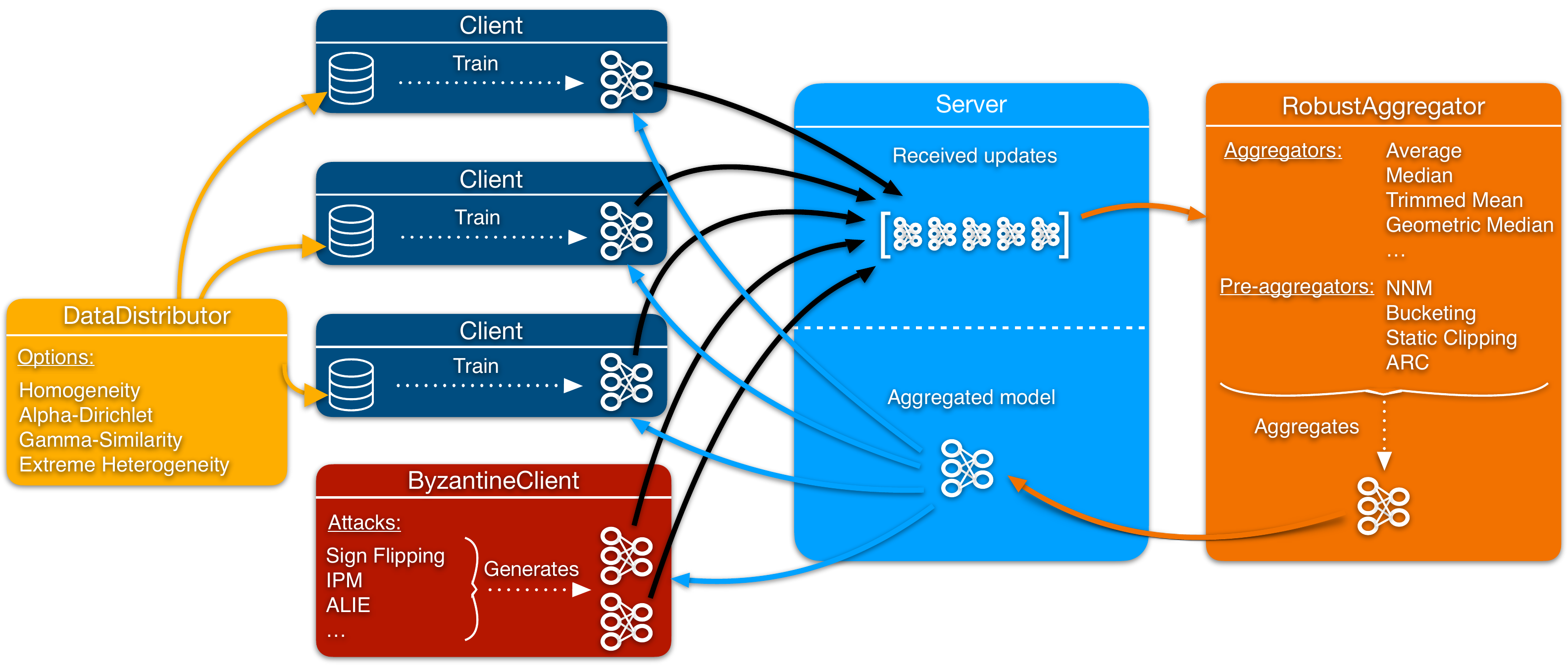}
    \caption{Architecture of a robust FL pipeline using \textsc{ByzFL}}
    \label{fig:diagram}
\end{figure}

\textsc{ByzFL} also includes a modular, user-friendly, and extensible FL simulation framework allowing researchers to flexibly configure clients (honest or Byzantine), define training protocols, and monitor learning performance.
At its core, the framework comprises four key components.
The \texttt{Client} class implements the behavior of honest clients, including local training, gradient computation, and update transmission. The \texttt{ByzantineClient} class simulates adversarial participants capable of executing a wide range of attack strategies, as discussed in Section~\ref{aggregators}. The \texttt{Server} class manages global coordination, including model aggregation and update broadcasting; it integrates with the \texttt{RobustAggregator} module, which implements the robust (pre-)aggregation schemes described in Section~\ref{aggregators}. Finally, the \texttt{DataDistributor} class is responsible for partitioning datasets among clients according to user-defined heterogeneity parameters, as detailed below.
The interaction between these core components is illustrated in the system diagram shown in Figure~\ref{fig:diagram}.

\paragraph{Simulating data heterogeneity.} 
Controlling the degree of data heterogeneity across clients is a key consideration in evaluating FL algorithms, particularly in the presence of adversaries. \textsc{ByzFL} provides a dedicated \texttt{DataDistributor} module to simulate both IID and non-IID client datasets.
For non-IID scenarios, \textsc{ByzFL} supports multiple partitioning schemes, including Dirichlet-based~\citep{dirichlet} and gamma-similarity-based~\citep{karimireddy2020scaffold} distributions. These methods allow fine-grained control over the degree of heterogeneity, enabling the simulation of realistic challenges such as label skew~\citep{zhao2018federated}. 
Such flexible data partitioning ensures that researchers can systematically assess the robustness of aggregation methods under a variety of realistic federated settings.
A code snippet showing the splitting of the MNIST dataset~\citep{mnist} over 5 honest clients using a Dirichlet distribution of parameter 0.5 is shown in Listing~\ref{lst:heterogeneity}.
\vspace{10pt}
\begin{lstlisting}[language=Python, caption=Splitting MNIST across 5 honest clients using a Dirichlet distribution, label={lst:heterogeneity}]
from torchvision.datasets import MNIST
from torchvision.transforms import ToTensor
from torch.utils.data import DataLoader
from byzfl import DataDistributor

# Load the MNIST dataset
dataset = MNIST(root="./data", train=True, transform=ToTensor())
# Define the parameters of the dataset distributor
params = {
    "data_distribution_name": "dirichlet_niid",
    "distribution_parameter": 0.5,
    "nb_honest": 5,
    "data_loader": DataLoader(dataset, shuffle=True),
    "batch_size": 64,
}

distributor = DataDistributor(params)
dataloaders = distributor.split_data()  # Returns per-client DataLoaders
\end{lstlisting}

\paragraph{Full FL benchmarking.}
Beyond offering modular simulation tools, \textsc{ByzFL} includes a dedicated \textit{FL Benchmark} module designed to facilitate easily configurable and reproducible experimentation in adversarial federated settings. This benchmarking suite orchestrates complete FL pipelines across a wide range of configurations, integrating honest clients, Byzantine adversaries, server logic, and robust aggregation strategies. \medskip

At its core, the benchmark operates through a single \texttt{config.json} file, which declaratively specifies the experiment's components: number of clients, data distribution schemes, attack strategies, robust aggregators, training hyperparameters, and evaluation metrics.
An example of the \texttt{config.json} file is presented in Listing~\ref{lst:config} in Appendix~\ref{app_code}.
This design minimizes user intervention and maximizes reproducibility, enabling the exhaustive testing of federated setups with minimal scripting.
Once configured, experiments are executed via a single call to the \texttt{run\_benchmark} function (see Listing~\ref{lst:run}), which automatically handles training execution, result storage, and metric aggregation.
The benchmark currently supports the two most widely adopted FL training paradigms, FedAvg and DSGD. These two algorithms were selected due to the availability of robustified variants in recent literature and their well-established theoretical guarantees under Byzantine settings~\citep{guerraoui2024robust,farhadkhani24FL}. By enabling the seamless integration of these training protocols with the robust (pre-)aggregators described in Section~\ref{aggregators}, \textsc{ByzFL} offers a principled and extensible platform for evaluating aggregation defenses in adversarial environments.

\begin{lstlisting}[language=Python, caption=Launching federated training experiments with the \textsc{ByzFL} benchmark, label={lst:run}]
from byzfl.benchmark import run_benchmark

if __name__ == "__main__":  # Required for multiprocessing
    n = 1  # Number of trainings to run in parallel
    run_benchmark(n)
\end{lstlisting}

\begin{figure}[ht!]
    \centering
    \includegraphics[width=0.49\linewidth]{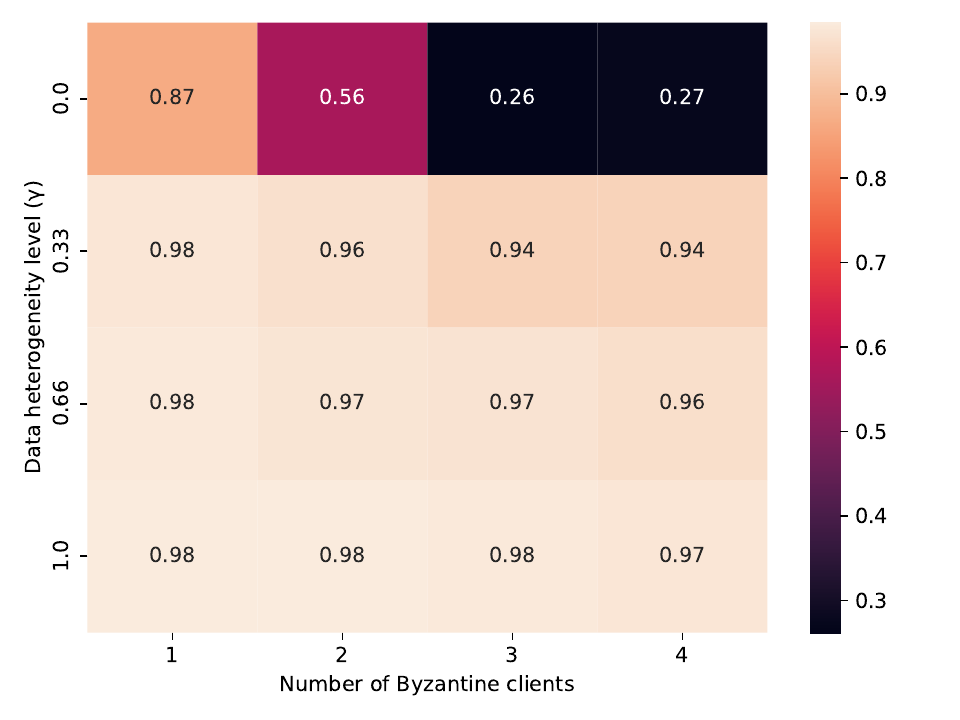}
    \includegraphics[width=0.49\linewidth]{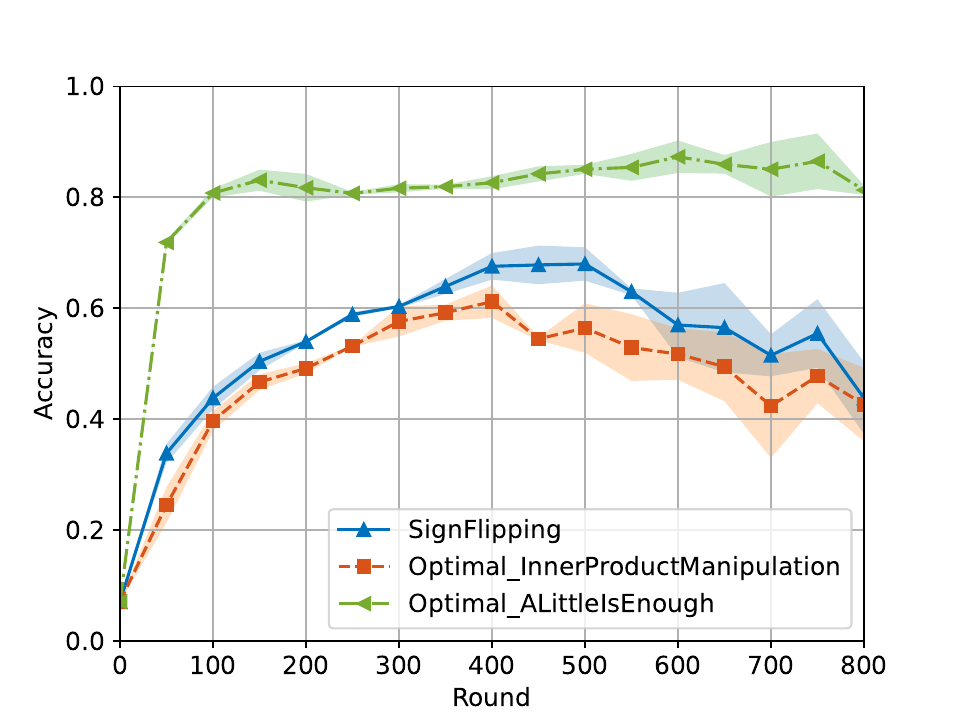}
    \vspace{-10pt}
    \caption{Performance of the trimmed mean aggregator~\citep{yin2018byzantine} under adversarial conditions with 10 honest clients. \textbf{Left:} Heatmap of test accuracy across varying numbers of Byzantine clients and gamma heterogeneity levels. \textbf{Right:} Test accuracy over training rounds for 3 attacks with 2 adversaries and highly heterogeneous data ($\gamma=0$).}
    \label{fig:plots}
\end{figure}

\paragraph{Result interpretation and visualization.}
To support the analysis of experimental outcomes, \textsc{ByzFL} includes a suite of built-in utilities for result visualization and interpretation. In particular, it enables the automatic plotting of test accuracy trajectories for each aggregator under a variety of attack scenarios, providing insight into convergence behavior and robustness over time. 
\textsc{ByzFL} also supports the generation of heatmaps that summarize performance metrics (e.g., test accuracy, training loss) across key experimental axes such as the number of Byzantine clients and the degree of data heterogeneity. These visual tools facilitate a systematic comparison of aggregation strategies under different federated settings.
\noindent Figure~\ref{fig:plots} presents an example output for the trimmed mean aggregator~\citep{yin2018byzantine} with clipping and NNM as pre-processing steps, in a federated setting executing DSGD with 10 honest clients. The left panel shows a heatmap of achieved test accuracies across a grid of adversarial and heterogeneity settings, while the right panel depicts the accuracy evolution over training rounds for three different attack strategies with $2$ Byzantine clients and highly heterogeneous data ($\gamma = 0$).
Listing~\ref{lst:results} illustrates the use of \textsc{ByzFL}’s high-level plotting API. By simply specifying the result and output directories, users can visualize experiment outcomes without needing to manually process logs or metrics.

\begin{lstlisting}[language=Python, caption=Generating accuracy curve and heatmap from training logs, label={lst:results}]
from byzfl.benchmark.evaluate_results import test_accuracy_curve, test_accuracy_heatmap

path_training_results = "./results"
path_plots = "./plot"
# Generate test accuracy curve and heatmap
test_accuracy_curve(path_training_results, path_plots)
test_accuracy_heatmap(path_training_results, path_plots)
\end{lstlisting}

\vspace{-5pt}

\paragraph{Worst-case performance metric.}
To assess the robustness of aggregators under comprehensive adversarial conditions, \textsc{ByzFL} introduces a \textit{worst-case maximal accuracy} metric, which is used to populate the cells of the generated heatmaps (see Figure~\ref{fig:plots}). For each configuration in the benchmark, defined by a fixed number of Byzantine clients and a given data heterogeneity level, \textsc{ByzFL} executes the full suite of attacks specified by the user in the configuration file. For each attack, it records the \textit{maximum test accuracy} achieved by the aggregator during training. The reported metric in the heatmap is then the \textit{minimum} of these maxima, representing the worst-case performance across the considered attacks. \medskip

This evaluation strategy is particularly important because different robust aggregation methods often exhibit highly variable behavior across attack types—sometimes performing well against specific threats they were designed to counter, while degrading under others. For instance, as shown in the right plot of Figure~\ref{fig:plots}, the trimmed mean aggregator~\citep{yin2018byzantine} performs well under the Opt-ALIE attack but suffers under the other two. The worst-case maximal accuracy provides an attack-agnostic measure of robustness, capturing how well an aggregator performs under the most damaging threat within the user-specified suite. This allows for a more meaningful and fair assessment of robustness, supporting rigorous comparisons between aggregators regardless of which attacks they may be implicitly designed to defend against.

\section{Conclusion}\label{conclusion}
We introduced \textsc{ByzFL}, an open-source Python library for benchmarking and prototyping robust FL algorithms. Through its modular architecture, support for robust aggregation and pre-aggregation, adversarial threat modeling, and systematic benchmarking, \textsc{ByzFL} enables reproducible and scalable experimentation in adversarial FL settings. As shown in Listings~\ref{lst:robg_agg} through~\ref{lst:results}, each core interaction with \textsc{ByzFL}, from executing robust aggregation mechanisms, to defining experimental configurations, launching training, and generating plots, can be performed in just a few lines of code. This low implementation overhead underscores the library’s emphasis on accessibility and ease of use. We expect \textsc{ByzFL} to have a meaningful impact on the research community by promoting fair and rigorous comparisons between robust aggregation methods, reducing the technical overhead of implementing new algorithms, and standardizing evaluation under realistic adversarial conditions. By lowering the barrier to entry for robust FL experimentation, \textsc{ByzFL} can accelerate progress in developing trustworthy, deployable FL systems. We hope it will serve as both a reproducible framework for future research and a reliable baseline implementation for practitioners.

\appendix
\section{Code Snippets}
\label{app_code}

\begin{lstlisting}[language=Python, basicstyle=\ttfamily\footnotesize, caption=Simple usage of \textsc{ByzFL} with \texttt{TrMean} aggregator and \texttt{SignFlipping} attack, label={lst:pipeline}]
import byzfl, torch

# Define honest gradients
honest_vectors = torch.tensor([[1., 2., 3.],
                               [4., 5., 6.],
                               [7., 8., 9.]])
# Define number of Byzantine clients
f = 1

# Apply Byzantine attack (Sign Flipping)
attack = byzfl.SignFlipping()
byz_vector = attack(honest_vectors)

# Repeat the attack vector f times to simulate f Byzantine clients
byz_vectors = byz_vector.repeat(f, 1)

# Combine honest and Byzantine vectors
all_vectors = torch.cat((honest_vectors, byz_vectors), dim=0)

# Robust aggregation using Trimmed Mean
aggregator = byzfl.TrMean(f=f)
print("Aggregated result:", aggregator(all_vectors))
# Output: Aggregated result: tensor([2.5000, 3.5000, 4.5000])
\end{lstlisting}

\clearpage
\begin{lstlisting}[language=json, 
  caption=Sample \texttt{config.json} file defining a robust FL experiment in \textsc{ByzFL}, 
  label={lst:config}]
{
    "benchmark_config": {
        // Option 1: Distributed SGD (DSGD)
        // "training_algorithm": {"name": "DSGD", "parameters": {}},

        // Option 2: Federated Averaging (FedAvg)
        "training_algorithm": {
            "name": "FedAvg",
            "parameters": {
                "proportion_selected_clients": 0.6,
                "local_steps_per_client": 5
            }
        },

        "nb_steps": 800,
        "nb_training_seeds": 3,
        "nb_honest_clients": 10,
        "f": [1, 2, 3, 4],
        "data_distribution": [
            {
                "name": "gamma_similarity_niid",
                "distribution_parameter": [1.0, 0.66, 0.33]
            }
        ]
    },

    "model": {
        "name": "cnn_mnist",
        "dataset_name": "mnist",
        "loss": "NLLLoss",
        "learning_rate": 0.1,
        "learning_rate_decay": 0.5,
        "milestones": [200, 400]
    },

    "aggregator": [
        {"name": "Median", "parameters": {}},
        {"name": "TrMean", "parameters": {}}
    ],

    "pre_aggregators": [
        {"name": "Clipping", "parameters": {"c": 2.0}}
    ],

    "honest_clients": {
        "momentum": 0.9,
        "weight_decay": 0.0001,
        "batch_size": 25
    },

    "attack": [
        {"name": "SignFlipping", "parameters": {}},
        {"name": "ALittleIsEnough", "parameters": {}}
    ],

    "evaluation_and_results": {
        "evaluation_delta": 50,
        "store_per_client_metrics": true,
        "results_directory": "./results"
    }
}
\end{lstlisting}

\vskip 0.2in
\bibliographystyle{plainnat}
\bibliography{ref}

\end{document}